\documentclass[11pt,a4paper]{article}
\usepackage{graphicx}
\usepackage{qtree}
\usepackage{acl2000}
\title{Tree-gram Parsing \\ Lexical Dependencies and Structural Relations}
\author{K. Sima'an
\\
        Induction of Linguistic Knowledge, Tilburg University \&\\
        Computational Linguistics, University of Amsterdam, \\
        Spuistraat 134, 1012 VB Amsterdam, The Netherlands.\\
        Email:~khalil.simaan@hum.uva.nl
}
\date{}
\input epsf
\newcommand{\Xsmall}{\small}

\newcommand{\lraN}{\rightarrow}

\newcommand{\RuleM}[2]{{#1\lraN #2}}

\newcommand{\TABL}[4]{
              \begin{tabular}{ll}
              #1 & #2 \\
              #3 & #4 \\
              \end{array}}

\begin{document}
\maketitle
\begin{abstract}
This paper explores the kinds of probabilistic relations that are important in
syntactic disambiguation. It proposes that two widely used kinds of relations,
{\em lexical dependencies}\/ and {\em structural relations}, have complementary 
disambiguation capabilities. 
It presents a new model based on structural relations, the {\em Tree-gram model}, 
and reports experiments 
showing that structural relations should benefit from enrichment by lexical dependencies.
\end{abstract}

\section{Introduction}
%
Head-lexicalization currently pervades in the parsing literature
e.g.~\cite{Eisner96,Collins97,Charniak99}.
This method extends every treebank nonterminal with its head-word:
the model is trained on this {\em head lexicalized treebank}. 
Head lexicalized models extract probabilistic relations between {\em pairs of lexicalized 
nonterminals} (``bilexical dependencies"): every relation is between a parent node and one 
of its children in a parse-tree. Bilexical dependencies generate parse-trees 
for input sentences via Markov processes that generate Context-Free Grammar (CFG) rules 
(hence Markov Grammar~\cite{Charniak99}). 

Relative to Stochastic CFGs (SCFGs), bilexical dependency models exhibit 
good performance. However, bilexical dependencies capture many but not all 
relations between words that are crucial for syntactic disambiguation.
We give three examples of kinds of relations not captured by bilexical-dependencies.
Firstly, relations between non-head words of 
phrases, e.g.\ the relation between ``more" and ``than" in 
``more apples than oranges" 
or problems of PP~attachments as in ``he ate pizza (with mushrooms)/(with a fork)".
Secondly, relations between three or more words are, by definition, beyond bilexical dependencies 
(e.g. between ``much more" and ``than" in ``much more apples than oranges"). 
Finally, it is unclear how bilexical dependencies help resolve the ambiguity of idioms, 
e.g.\ ``Time flies like an arrow" (neither ``time" prefers to ``fly", 
nor the fictitios beasts ``Time flies" have taste for an ``arrow"). 

The question that imposes itself is, indeed, {\em what relations might 
complement bilexical dependencies~?}
We propose that bilexical dependencies can be complemented by 
{\em structural relations} \cite{Scha90}, i.e.\ cooccurrences of syntactic structures, including 
actual words. An example model that employs one version of structural relations is Data Oriented 
Parsing (DOP)~\cite{RENSDES}. 
DOP's parameters are ``subtrees", i.e.\ connected subgraphs of parse-trees that constitute 
combinations of CFG rules, including terminal rules. 

Formally speaking, ``bilexical dependencies" and ``structural relations" define two 
disjoint sets of probabilistic relations. Bilexical dependencies are relations defined 
over direct dominance head lexicalized nonterminals (see~\cite{Satta2000}); 
in contrast, structural relations are defined over words and arbitrary size 
syntactic structures (with non-lexicalized nonterminals).
Apart from formal differences, they also have complementary advantages.
Bilexical-dependencies capture influential lexical relations between heads and dependents.
Hence, all bilexical dependency probabilities are {\em conditioned on lexical information}\/
and lexical information is available at every point in the parse-tree.
Structural relations, in contrast, capture many relations not captured by bilexical-dependencies (e.g. the examples above).
However, structural relations do not always percolate lexical information up the parse-tree
since their probabilities are not always {\em lexicalized}. This is a serious 
disadvantage when parse-trees are generated for {\em novel}\/ input sentences since e.g.\ 
subcat frames are hypothesized for nodes high in the parse-tree without reference 
to their head words. 

So, theoretically speaking, bilexical dependencies and structural relations have
complementary aspects. But, {\em what are the empirical merits and limitations of structural 
relations~?}
%
This paper presents a new model based on {\em structural relations}, 
the Tree-gram model, which allows head-driven parsing. It studies the effect of
percolating head categories on performance and compares the performance of structural
relations to bilexical dependencies. The comparison is 
conducted on the Wall Street Journal (WSJ) corpus~\cite{Santorini}. 
%
In the remainder, we introduce the Tree-gram model in section~\ref{tgrams},
discuss practical issues in section~\ref{Practical}, 
exhibit and discuss the results in section~\ref{EmpiricSec}, 
and in section~\ref{SecConcs} we give our conclusions.

\newcommand{\VSP}[1]{^{#1}}
\newcommand{\SIZE}[1]{{\Large #1}}
\newcommand{\E}{${\mathcal C}$}
\newcommand{\EN}{{\mathcal C}}
\newcommand{\Hcal}{${\mathcal H}$}
\newcommand{\Lcal}{${\mathcal L}$}
\newcommand{\Rcal}{${\mathcal R}$}
\newcommand{\HcalN}{{\mathcal H}}
\newcommand{\LcalN}{{\mathcal L}}
\newcommand{\RcalN}{{\mathcal R}}
\newcommand{\ENPi}{\EN_{\Pi}}
\newcommand{\sdelta}[2]{{\delta_{#1}^{#2}}}
\newcommand{\OnTop}[3]{\fbox{$\stackrel{{\tiny #3}}{{{\tiny #1}}^{\tiny #2}}$}}
\section{The Tree-gram model}
\label{tgrams}

For observing the effect of percolating information up the parse-tree on model behavior,
we introduce {\em pre-head enrichment}, a structural variant of head-lexicalization. 
Given a training treebank~$TB$, for every non-leaf node $\mu$ we mark one of its children as 
the {\em head-child}, i.e.\ the child that dominates the head-word\footnote{ 
Head-identification procedure by~\cite{Collins97}.} of the constituent under $\mu$. 
%
%
We then enrich this treebank by attaching to the label of every phrasal node
(i.e.\ nonterminal that is not a POS-tag) a {\em pre-head}\/ representing its head-word. 
The pre-head of node~$\mu$ is extracted from the constituent parse-tree under node~$\mu$.
In this paper, the pre-head of~$\mu$ consists of 1)~the POS-tag of the head-word of ~$\mu$ 
(called $1^{st}$ order pre-heads or $1^{PH}$), and possibly
2)~the label of the mother node of that POS-tag (called $2^{nd}$ order or $2^{PH}$). 
Pre-heads here also include other information defined in the sequel, e.g.\ subcat frames. 
The complex categories that result from the enrichment serve as the 
nonterminals of our training treebank; we refer to the original treebank symbols as ``WSJ~labels".
\subsection{Generative models}
A probabilistic model assigns a probability to every parse-tree given an input sentence $S$,
thereby distinguishing one parse
%
 \mbox{$T^* = arg max_{T}~ P(T | S) =  arg max_{T}~ P(T,S)$.}
%
The probability $P(T,S)$ is usually estimated from cooccurrence statistics 
extracted from a treebank. In generative models, the tree $T$ is generated in top down 
derivations that rewrite the start symbol $TOP$ into the sentence $S$. Each rewrite-step
involves a ``rewrite-rule" together with its estimated probability.
%
In the present model, the ``rewrite-rules" differ from the CFG rules and combinations thereof 
that can be extracted from the treebank. We refer to them as {\em Tree-grams}\/ 
(abbreviated T-grams).  T-grams provide a more general-form for 
Markov Grammar rules~\cite{Collins97,Charniak99} as well as DOP subtrees.  In comparison
with DOP subtrees, T-grams capture more structural relations, allow head-driven parsing and
are easier to combine with bilexical-dependencies.
%
\subsection{T-gram extraction}
Given a parse~$T$ from the training treebank, we extract three disjoint T-gram sets, 
called {\em roles}, from every one of its non-leaf nodes\footnote{Assuming that every node has a unique address.}~$\mu$: 
the head-role \Hcal$(\mu)$, the left-dependent role \Lcal$(\mu)$ and the right-dependent
role \Rcal$(\mu)$. 
The role of a T-gram signifies the T-gram's contribution to stochastic derivations: 
$t\in\HcalN$\ carries a head-child of its root node label, $t\in\LcalN$\ ($t\in\RcalN$) 
carries left (resp.\/ right) dependents for other head T-grams that have roots labeled the 
same as the root of~$t$. Like in Markov Grammars, a head-driven derivation generates first a 
head-role T-gram and attaches to it left- and right-dependent role T-grams. We discuss
these derivations right after we specify the T-gram extraction procedure.
\begin{figure}[tb]
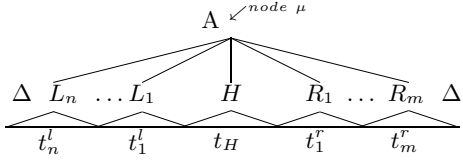

\center{
\parbox[t]{3.2in}{
{\footnotesize
\Tree  [.{~~~~~~A$^{~\swarrow\VSP{node~\mu}}$}
                \qroof{~~~$t_n^l$~~~~~}.{~~~~$\Delta~~L_n$~~\ldots} !\faketreewidth{WWl}   
                \qroof{~~~$t_1^l$~~~~~}.{$L_1$}  !\faketreewidth{Wl}
                \qroof{~~~$t_H$~~~~~}.{$H$}                       !\faketreewidth{WWl}
                \qroof{~~~$t_1^r$~~~~~}.{~~~~~$R_1~\ldots$}   !\faketreewidth{WWl}
                \qroof{~~~$t_m^r$~~~~~}.{~~~~$R_m~~\Delta$}  !\faketreewidth{WWl}
       ]
}
}
}
\caption{Constituent under node $\mu$: $d >1$.}
\label{FigRuleF}
\end{figure}

%
Let $d$ represent the depth\footnote{The depth of a (sub)tree is the number of 
edges in the longest path from its root to a leaf node.} of the constituent tree-structure that is
rooted at~$\mu$, $H$ represent the label of the head-child of $\mu$, and $\Delta$ represent
the special {\em stop}\/ symbol that encloses the children of every node (see figure~\ref{FigRuleF}). 
Also, for convenience, let $\sdelta{k}{n}$ be equal to $\Delta$ iff $k=n$ and $NILL$ (i.e. the empty tree-structure) 
otherwise. We specify the extraction for \mbox{$d = 1$} and for \mbox{$d > 1$}.
When $d=1$, the label of~$\mu$ is a POS-tag and the subtree under $\mu$ is of the form $\RuleM{pt}{\Delta w\Delta}$, 
where $w$ is a word. In this case \Hcal$(\mu)=\{\RuleM{pt}{\Delta w\Delta}\}$ and 
\Lcal$(\mu)$ = \Rcal$(\mu)$ = $\emptyset$.
When $d > 1$: the subtree under $\mu$ has the form 
{ 
\Xsmall
$\RuleM{A}{\Delta L_{n}(t_n^l)\ldots L_{1}(t_1^l)~H(t_H)~R_{1}(t_1^r)\ldots R_m(t_m^r)\Delta}$
} (figure~\ref{FigRuleF}), where every $t_i^l$, $t_j^r$ and $t_H$ is the subtree dominated 
by the child node of $\mu$ (labeled respectively $L_i$, $R_j$ or $H$) 
whose address we denote respectively with $child_L(\mu,i)$, $child_R(\mu,j)$ and 
$child_H(\mu)$. We extract three sets of T-grams from $\mu$:
\begin{description}
\item [\Hcal$(\mu)$ :] contains \mbox{$\forall\  1\leq i < n$ and $1\leq j < m$,} 
   $\RuleM{A}{\sdelta{i}{n} L_{i}(X_i^l)\ldots H(X_h)\ldots R_{j}(X_j^r) \sdelta{j}{m}}$,
   %
    where  $X_h$ is either in 
    \Hcal$(child_H(\mu))$ or $NILL$, and every $X_z^l$ (resp.\/ $X_z^r$) is either a T-gram 
    from \Hcal$(child_L(\mu,z))$ (resp.\/ \Hcal$(child_R(\mu,z))$ ) or $NILL$.
\item [\Lcal$(\mu)$:] contains~ 
     \mbox{$\RuleM{A}{\sdelta{k}{n} L_{k}(X_k)\ldots L_{i}(X_i)}$},
     for all \mbox{$1\leq i\leq k < n$}, 
     where every $X_z$, \mbox{$i\leq z\leq k$}, is either a T-gram from 
     \Hcal$(child_L(\mu,z))$ or $NILL$, 
\item [\Rcal$(\mu)$ :] contains~ 
     \mbox{$\RuleM{A}{R_{i}(X_i)\ldots R_{k}(X_k)}\sdelta{k}{m}$}, 
     for all \mbox{$1\leq i\leq k < m$}, 
     where every $X_z$,  \mbox{$i\leq z\leq k$}, is either a T-gram from 
     \Hcal$(child_R(\mu,z))$ or $NILL$,
\end{description}
Note that every T-gram's non-root and non-leaf node dominates a {\em head-role}\/ T-gram 
(specified by~\Hcal$(child\cdots)$). 

A non-leaf node $\mu$ labeled by nonterminal $A$ is called {\em complete}, denoted ``$[A]$", 
iff $\Delta$ delimits its sequence of children from both sides; when $\Delta$ is to the left 
(right) of the children of the node, the node is called {\em left (resp.\/ right) complete}, 
denoted ``$[A$" (resp.\/ ``$A]$"). When $\mu$ is not left (right) complete it is 
{\em open from the left}\/ (resp. {\em right}); when~$\mu$ is left and right open, it is 
called {\em open}.
\begin{figure}[tb]
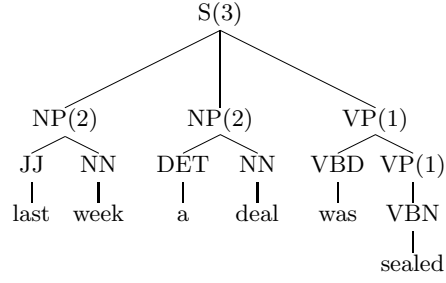

\parbox[t]{2.83in}{
{\footnotesize
\Tree [.S(3)
         [.{NP(2)}  [.{JJ} last ] [.{NN} week ]   !\faketreewidth{Wl}   ]
         [.NP(2)  [.{DET} a ] [.{NN} deal    ]    !\faketreewidth{Wl} ]
         [.{VP(1)}   [.{VBD} was ]  [.{VP(1)}  [.VBN sealed ] !\faketreewidth{Wl} ]  !\faketreewidth{Wl}  ] ] 
}
}
\caption{An example  parse-tree. 
        }
\label{FigExample1}
\end{figure}

\begin{figure*}[tb]
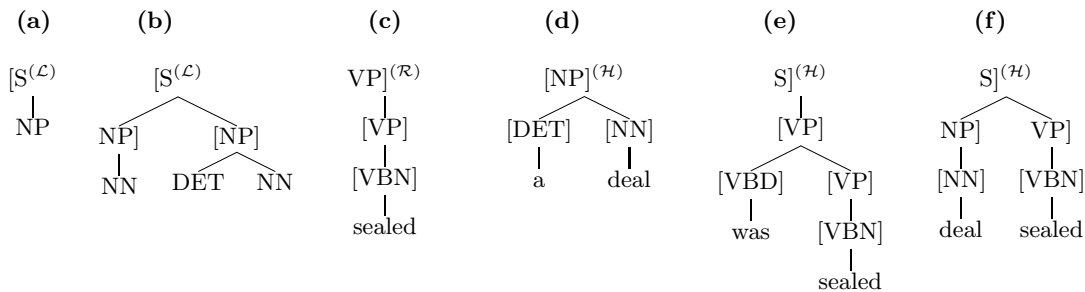

\mbox{
{\footnotesize
\parbox[t]{1.1cm}{ \hskip 1em {\bf(a)\vspace*{1.0em}}\Tree [.{{[S}$^{(\LcalN)}$} NP ] }
\parbox[t]{3.2cm}{ \hskip 2.2em{\bf(b)\vspace*{1.0em}}\Tree [.{{[S}$^{(\LcalN)}$} [.{NP]} NN ] [.{[NP]} DET NN ] ] }
\parbox[t]{2.0cm}{ \hskip 1.5em{\bf(c)\vspace*{1.0em}}\Tree [.{{VP]}$^{(\RcalN)}$} [.{[VP]} [.{[VBN]} sealed ] ] ] }
\parbox[t]{2.7cm}{ \hskip 2.2em{\bf(d)\vspace*{1.0em}}\Tree [.{[NP]$^{(\HcalN)}$} [.{[DET]} a ] [.{[NN]} deal ] ] }
\parbox[t]{2.8cm}{ \hskip 2.5em{\bf(e)\vspace*{1.0em}}\Tree [.{{S]}$^{(\HcalN)}$} [.{[VP]} [.{[VBD]} was ] [.{[VP]} [.{[VBN]} sealed ]  ] ] ] }
\parbox[t]{2.7cm}{ \hskip 2.2em{\bf(f)\vspace*{1.0em}}\Tree [.{{S]}$^{(\HcalN)}$} [.{NP]} [.{[NN]} deal ] ] [.{VP]} [.{[VBN]} sealed ] ] ] }
}
}
\caption{{ 
\Xsmall Some T-grams extracted from the tree in figure~\ref{FigExample1}: the superscript on the root label 
         specifies the {\em T-gram role},. e.g.\ the left-most T-gram is in the left-dependent role.
         Non-leaf nodes are marked with ``[" and ``]" to specify whether they are complete from the left/right 
         or both (leaving open nodes unmarked).
        }}
\label{FigExample2}
\end{figure*}

Figure~\ref{FigExample1} exhibits a parse-tree\footnote{Pre-heads are omitted for readability.}:
the number of the head-child of a node is specified between brackets. 
Figure~\ref{FigExample2} shows some of the T-grams that can be extracted from this tree.

Having extracted T-grams from all non-leaf nodes of the treebank, we obtain
\mbox{\Hcal$ = \bigcup_{\mu\in TB}$\Hcal$(\mu)$}, \mbox{\Lcal$ = \bigcup_{\mu\in TB}$\Lcal$(\mu)$}
 and \mbox{\Rcal$ = \bigcup_{\mu\in TB}$\Rcal$(\mu)$}. 
$\HcalN_A$, $\LcalN_A$ and $\RcalN_A$ represent the subsets of 
resp.\/ \Hcal, \Lcal\  and \Rcal\  that contain those T-grams that have roots labeled~$A$. 
\mbox{$X_A(B)\in \{\LcalN_A(B), \RcalN_A(B), \HcalN_A(B)\}$} specifies
that the extraction took place on some treebank $B$ other than the training treebank.
%
%
%
\subsection{T-gram generative processes}
Now we specify T-gram derivations assuming that we have an estimate of the probability of a T-gram.
We return to this issue right after this. A stochastic derivation starts from the start nonterminal $TOP$. 
$TOP$ is a single node partial parse-tree which is simultaneously the root and the only leaf node.
A derivation terminates when two conditions are met 
(1)~every non-leaf node in the generated parse-tree is {\em complete}\/ (i.e. $\Delta$~delimits 
its children from both sides) and 
(2)~all leaf nodes are labeled with terminal symbols.
%
Let $\Pi$ represent the current partial parse-tree, i.e.\/ the result of the 
preceding generation steps, and let $\ENPi$ represent that part of~$\Pi$ 
that influences the choice of the next step, i.e.\/ the conditioning history. 
The generation process repeats the following steps in some order, e.g.\ head-left-right:
%
\paragraph{\bf Head-generation:}Select a leaf node~$\mu$ labeled by a nonterminal~$A$, and let~$A$ 
      generate a head T-gram~$t\in\HcalN_A$ with probability~$P_H(t | A, \ENPi)$. This results in 
      a partial parse-tree that extends~$\Pi$ at~$\mu$  with a copy of~$t$ (as in CFGs and in~DOP).
\paragraph{\bf Modification:}Select from $\Pi$ a non-leaf node~$\mu$ that is {\em not complete}.
      Let $A$ be the label of~$\mu$ and~$T = \RuleM{A}{X_1(x_1)\cdots X_b(x_b)}$ be the 
      tree dominated by~$\mu$ (see figure~\ref{FigLRMod}):
 \begin{description}
  \item [Left:] if $\mu$ is not left-complete, let $\mu$ generate to the left of $T$ a left-dependent T-gram
      \mbox{$t = \RuleM{A^{(\LcalN)}}{L_1(l_1)\cdots L_a(l_a)}$} from $\LcalN_A$ with probability $P_L(t | A, \ENPi)$ 
      (see figure~\ref{FigLRMod}~(L)); this results in a partial parse-tree that is obtained by 
      replacing $T$ in $\Pi$ with $\RuleM{A}{L_1(l_1)\cdots L_a(l_a) X_1(x_1)\cdots X_b(x_b)}$, 
  \item [Right:] this is the mirror case (see figure~~\ref{FigLRMod}~(R)). The 
      generation probability is $P_R(t | A, \ENPi)$.
 \end{description}
Figure~\ref{FigExample3} shows a derivation using T-grams (e), (a) and (d) from figure~\ref{FigExample2}
applied to T-gram~\mbox{$TOP\lraN S$}.
\begin{figure*}[tb]
{ 
\begin{tabular}{llp{0.01cm}lp{0.2cm}l}
\parbox[t]{0.001in}{ (L) } &
\parbox{1.1in}{
{\footnotesize
\Tree  [.{$A^{(\LcalN)}$}
                \qroof{~~~~~~~~~~}.{~~~~~~$L_1$~~~\ldots} !\faketreewidth{WW}   
                \qroof{~~~~~~~~~~}.{$L_a$} ]
}
}
& \hspace*{-1.3em}\begin{tabular}{c} \\ {\large $+$~~~~}\\ \\\\\end{tabular}& \hspace*{-3.0em}
\parbox{1.1in}{
{\footnotesize
\Tree  [.{\vdots\\ ~~~~~~~~$A^{~\swarrow\VSP{node~\mu}}$}
                \qroof{~~~~~~~~~~}.{~~~~~~$X_1$~~~\ldots} !\faketreewidth{WW}   
                \qroof{~~~~~~~~~~}.{$X_b$} ] 
}
}
&
{\Large ~~$\Longrightarrow$}
& \hspace*{-0.3em}
\parbox{1in}{
{\footnotesize
\Tree  
             [.{\vdots\\ ~~~~~~~~~~~~~~~~$A^{~\swarrow\VSP{node~\mu~after}}$} !\faketreewidth{}
                \qroof{~~~~~~~~~~}.{~~~~~~$L_1$~~~\ldots} !\faketreewidth{WW} 
                \qroof{~~~~~~~~~~}.{$L_a$}
                \qroof{~~~~~~~~~~}.{~~~~~~$X_1$~~~\ldots} !\faketreewidth{WW} 
                \qroof{~~~~~~~~~~}.{$X_b$} 
             ]
}
}
\\
\parbox{0.001in}{ (R) } &
\parbox{1.1in}{
{\footnotesize
\Tree [.{\vdots\\ ~~~~~~~~~~$A^{~\swarrow\VSP{node~\mu}}$} 
                \qroof{~~~~~~~~~~}.{~~~~~~$X_1$~~~\ldots} !\faketreewidth{WW}   
                \qroof{~~~~~~~~~~}.{$X_b$} ] 
}
}
& \hspace*{-1.3em} \begin{tabular}{c} \\ {\large $+$}\\ \\\\\end{tabular} & \hspace*{-2.8em} 
\parbox{0.9in}{
{\footnotesize
\Tree  [.{$A^{(\RcalN)}$} 
                \qroof{~~~~~~~~~~}.{~~~~~~$R_1$~~~\ldots} !\faketreewidth{WW}   
                \qroof{~~~~~~~~~~}.{$R_a$} ]
}
}
&
{\Large ~~$\Longrightarrow$}
& \hspace*{-0.3em}
\parbox{1in}{
{\footnotesize
\Tree [.{\vdots\\ ~~~~~~~~~~~~~~~~$A^{~\swarrow\VSP{node~\mu~after}}$} 
                \qroof{~~~~~~~~~~}.{~~~~~~$X_1$~~~\ldots}   !\faketreewidth{WW} 
                \qroof{~~~~~~~~~~}.{$X_b$} 
                \qroof{~~~~~~~~~~}.{~~~~~~$R_1$~~~\ldots} !\faketreewidth{WW}
                \qroof{~~~~~~~~~~}.{$R_a$}
      ] 
}
}
\end{tabular}
}
\caption{T-gram $t$ is generated at $\mu$:~~(L)~$t\in\LcalN_A$,~~(R)~$t\in\RcalN_A$} 
\label{FigLRMod}
\end{figure*}
Note that each derivation-step probability is conditioned on~$A$, the label of node~$\mu$ in~$\Pi$ 
where the current rewriting is taking place, on the role (\Hcal, \Lcal\   or \Rcal) of the T-gram 
involved, and on the relevant history~$\ENPi$. 
Assuming beyond this that stochastic independence between the various 
derivation steps holds, {\em the probability of a derivation is equal to the 
multiplication of the conditional probabilities of the individual rewrite steps}.

%
{\tiny
\begin{figure*}[tb]
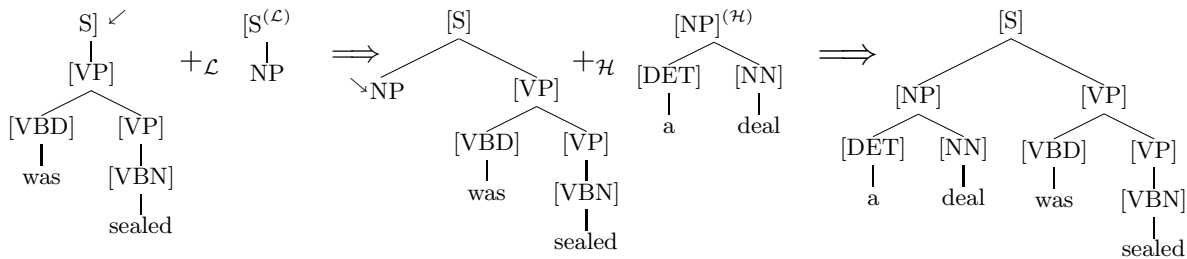

\mbox{
{\footnotesize
\parbox[t]{2.2cm}{ \Tree [.{~~~S]~$^\swarrow$} [.{[VP]} [.{[VBD]} was ] [.{[VP]} [.{[VBN]} sealed ]  ] ] ] }
\parbox[t]{0.7cm}{\vspace*{2em}{\large ~~$+_{\LcalN}$}}
\parbox[t]{1.1cm}{ \Tree [.{{[S}$^{(\LcalN)}$} NP ] }
\parbox[t]{0.0cm}{\vspace*{2em}{\large ~~$\Longrightarrow$}}
\parbox[t]{2.8cm}{ \Tree [.{[S]} ~$^\searrow$NP~~   [.{[VP]} [.{[VBD]} was ] [.{[VP]} [.{[VBN]} sealed ]  ] ] ] }
\parbox[t]{0.9cm}{\vspace*{2em}{\large ~~~~$+_{\HcalN}$}}
\parbox[t]{2.3cm}{ \Tree [.{[NP]$^{(\HcalN)}$} [.{[DET]} a ] [.{[NN]} deal ] ] }
\parbox[t]{0.2cm}{\vspace*{2em}{\Large ~~$\Longrightarrow$}}
\parbox[t]{1.5cm}{ \Tree [.{[S]} [.{[NP]} [.{[DET]} a ] [.{[NN]} deal ] ] [.{[VP]} [.{[VBD]} was ] [.{[VP]} [.{[VBN]} 
 sealed ]  ] ] ]  }
}
}
\caption{{ 
\Xsmall A T-gram derivation: the rewriting of TOP is not shown. An arrow marks the node where 
rewriting takes place. Following the arrows: 1.~A left T-gram with 
root $[S$ is generated at node $S]$: $S$ is complete. 2.~A head-role T-gram
is generated at node $NP$: all nodes are either complete or labeled with terminals.}}
\label{FigExample3}
\end{figure*}
}
%

Unlike SCFGs and Markov grammars but like DOP, a parse-tree may be generated 
via different derivations. The probability of a parse-tree~$T$ is equal to the 
{\em sum of the probabilities of the derivations that generate it 
(denoted \mbox{$der\Rightarrow T$})}, i.e.\ \mbox{$P(T,S) = \sum_{der\Rightarrow T} P(der,S)$}.
However, because computing \mbox{$arg max_{T}~ P(T,S)$}
can not be achieved in deterministic polynomial time \cite{MyCOLING96}, we apply estimation 
methods that allow tractable parsing. 
%

\newcommand{\LA}{\langle}
\newcommand{\RA}{\rangle}
\newcommand{\tuple}[1]{\LA #1\RA}

\subsection{Estimating T-gram probabilities}
%
Let $count(Y_1,\cdots Y_m)$ represent the occurrence count for joint event \mbox{$\LA Y_1\,\cdots Y_m\RA$} in the training treebank.
Consider a T-gram $t\in X_A$, \mbox{$X_A\in \{\LcalN_A, \RcalN_A, \HcalN_A\}$}, and a 
conditioning history $\ENPi$. 
The estimate $\frac{count(t, X_A, \ENPi)}{\sum_{x\in X_A} count(x, X_A, \ENPi)}$ assumes no 
hidden elements (different derivations per parse-tree), i.e.\ it estimates  the probability
\mbox{$P_X(t | A, \ENPi)$} directly from the treebank trees (henceforth {\em direct-estimate}). This estimate 
is employed in DOP and is not Maximum-Likelihood~\cite{BonnemaBuyingScha2000}.
We argue that the bias of the direct estimate allows approximating the preferred 
parse by the one generated by the Most Probable Derivation (MPD). This is beyond 
the scope of this paper and will be discussed elsewhere. 
\subsection{WSJ model instance} 
\label{WSJinstance}
\begin{figure*}[htb]
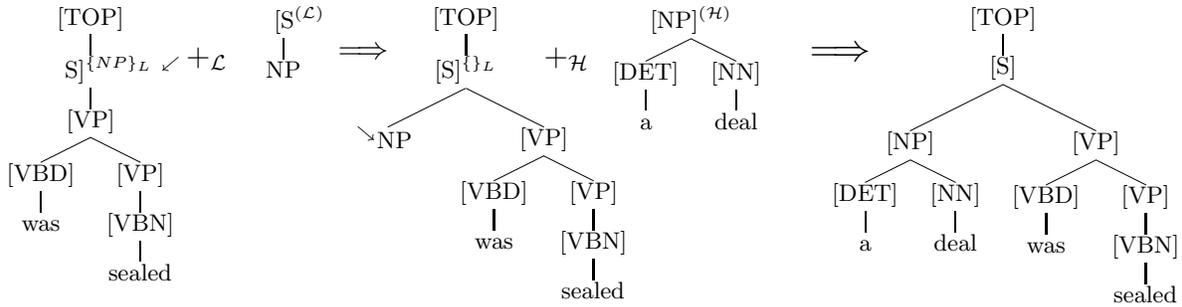

\mbox{
{\footnotesize
\parbox[t]{2.4cm}{ \Tree [.{[TOP]}  [.{~~~~~~~~S]$^{\{NP\}_L}$~$^\swarrow$} [.{[VP]} [.{[VBD]} was ] [.{[VP]} [.{[VBN]} sealed ]  ] ] ] ] }
\parbox[t]{0.6cm}{\vspace*{2em}{\large ~~$+_{\LcalN}$}}
\parbox[t]{1.2cm}{ \Tree [.{~~~~{[S}$^{(\LcalN)}$} NP ] }
\parbox[t]{0.0cm}{\vspace*{2em}{\large ~~$\Longrightarrow$}}
\parbox[t]{2.5cm}{ \Tree [.{[TOP]}  [.{[S]$^{\{\}_L}$} ~$^\searrow$NP~~   [.{[VP]} [.{[VBD]} was ] [.{[VP]} [.{[VBN]} sealed ]  ] ] ] ] }
\parbox[t]{0.8cm}{\vspace*{2em}{\large ~~$+_{\HcalN}$}}
\parbox[t]{2.5cm}{ \Tree [.{[NP]$^{(\HcalN)}$} [.{[DET]} a ] [.{[NN]} deal ] ] }
\parbox[t]{0.2cm}{\vspace*{2em}{\Large ~~$\Longrightarrow$}}
\parbox[t]{1.5cm}{ \Tree [.{[TOP]} [.{[S]} [.{[NP]} [.{[DET]} a ] [.{[NN]} deal ] ] [.{[VP]} [.{[VBD]} was ] [.{[VP]} [.{[VBN]} 
 sealed ]  ] ] ]  ] }
}
}
\caption{{ 
          \Xsmall $S]^{\{NP\}_L}$ is a (left-open right-complete) node labeled $S$ with a left subcat frame containing an NP. 
After the first rewriting, the subcat frame becomes empty since the NP complement was generated resulting in {[S]$^{\{\}_L}$}. 
The Other subcat frames are empty and are not shown here.}}
\label{FigExample4}
\end{figure*}

Up till now $\ENPi$ represented conditioning information anonymously in our model.
For the WSJ corpus, we instantiate $\ENPi$ as follows: 
%
\mbox{\sl {\bf 1.}~Adjacency:}
The flag~$F_{L}(t)$ ($F_R(t)$) tells whether a left-dependent (right-dependent) T-gram~$t$
extracted from some node~$\mu$ dominates a surface string that is adjacent to the
head-word of~$\mu$ (detail in~\cite{Collins97}). 
%
\mbox{\sl {\bf 2.}~Subcat-frames:}
\cite{Collins97} subcat frames are adapted: with every node~$\mu$ that dominates a rule
\mbox{$\RuleM{A}{\Delta L_{n}\ldots L_{1}~H~R_{1}\ldots R_m\Delta}$} in the treebank (figure~\ref{FigRuleF}),
we associate two (possibly empty) multisets of complements: ~$SC^{\mu}_L$ and 
~$SC^{\mu}_R$. Every complement in $SC^{\mu}_L$  ($SC^{\mu}_R$) represents some left (right)
complement-child of~$\mu$. 
This changes T-gram extraction as follows: {\em with every non-leaf node in a T-gram}\/ 
that is extracted from a tree in this enriched treebank we have now a left and a right 
subcat frame associated. Consider the root node~$x$ in a T-gram extracted from node~$\mu$ and let the
children of~$x$ be \mbox{$Y_1\cdots Y_f$} (a subsequence of
\mbox{$\Delta L_n,\cdots, H,\cdots R_m\Delta$}). The left (right) subcat frame of~$x$ is 
subsumed by $SC^{\mu}_L$ (resp. $SC^{\mu}_R$) and contains those complements
that correspond to the left-dependent (resp. right-dependent) children of~$\mu$ that are 
{\em not}\/ among~\mbox{$Y_1\cdots Y_f$}.
Tree-gram derivations are modified accordingly: whenever a T-gram is generated (together with
the subcat frames of its nodes) from some node~$\mu$ in a partial-tree, the complements that 
its root dominates are removed from the subcat frames of~$\mu$. Figure~\ref{FigExample4} shows a small
example of a derivation.
%
\mbox{\sl {\bf 3.}~Markovian generation:} When node $\mu$ has empty subcat frames, we assume $1st$-order
Markov processes in generating both $\LcalN$  and $\RcalN$ T-grams around its $\HcalN$~ T-gram: $LM^{\mu}$ and $RM^{\mu}$
denote resp.\ the left- and right-most children of node $\mu$. Let $XRM^{\mu}$ and $XLM^{\mu}$ be equal to resp.\ 
$RM^{\mu}$ and $LM^{\mu}$ if the name of the T-gram system contains the word $+Markov$ (otherwise they are empty).

Let~$\mu$, labeled $A$, be the node where the current rewrite-step takes place, 
$P$~be the WSJ-label of the parent of~$\mu$, and $H$~the WSJ-label of the head-child 
of~$\mu$. Our probabilities are:
{
\mbox{$P_H(t | A, \ENPi) \approx P_H(t | A, P)$,} 
\mbox{$P_L(t | A, \ENPi) \approx P_L(t | A, H, SC^{\mu}_L, F_L(t), XRM^{\mu})$,} 
{\mbox{$P_R(t | A, \ENPi) \approx P_R(t | A, H, SC^{\mu}_R, F_R(t), XLM^{\mu})$.}}
}
%
%

%
%
%
\newcommand{\PARA}[1]{\mbox{\underline{\sl #1:}}}
\section{Implementation issues}
\label{Practical}
Sections~02-21 WSJ Penn Treebank~\cite{Santorini} (release~2)
are used for training and section~23 is held-out for testing (we tune on section~24). 
The parser-output is evaluated by ``evalb"\footnote{http://www.research.att.com/~mcollins/.}, 
on the PARSEVAL measures~\cite{PARSEVAL} comparing a proposed parse~$P$ with
the corresponding treebank parse~$T$ on Labeled Recall 
({LR} =  $\frac{number~of~correct~constituents~in~P}{number~of~constituents~in~T}$),
Labeled Precision ({LP}  =  $\frac{number~of~correct~constituents~in~P}{number~of~constituents~in~P}$),
and Crossing Brackets ({CB}  = {number of constituents in P that violate constituent boundaries in T}).
%
%

%
\PARA{T-gram extraction}
The number of T-grams is limited by setting constraints on their form much like $n$-grams.
One upperbound is set on the depth\footnote{T-gram depth is the length of the longest path in
the tree obtained by right/left-linearization of the T-gram around the T-gram nodes' head-children.
} ($d$), a second on the number of children of every
node~($b$), a third on the sum of the number of nonterminal leafs with 
the number of (left/right) open-nodes ($n$), and a fourth ($w$) on the number of words in a
T-gram. Also, a threshold is set on the frequency ($f$) of the T-gram. 
In the experiments $n\leq4$, $w\leq3$ and $f\geq5$ are fixed while $d$ changes.
\PARA{Unknown words and smoothing}
We did not smooth the relative frequencies. Similar to~\cite{Collins97}, every word occurring less 
than~5 times in the training-set was renamed to CAP+UNKNOWN+SUFF, where CAP is~1 if its 
first-letter is capitalized and 0~otherwise, and SUFF is its suffix. 
Unknown words in the input are renamed this way before parsing starts.
\PARA{Tagging and parsing}
An input word is tagged with all POS-tags with which it cooccurred in the training treebank.
The parser is a two-pass CKY~parser: the first pass employs T-grams that fulfill $d=1$ in 
order to keep the parse-space under control before the second-pass employs the full 
Tree-gram model for selecting the~MPD. 
%
%
\section{Empirical results}
\label{EmpiricSec}
{ 
\Xsmall
\begin{table*}[tbh]
\center{
\begin{tabular}{|l|lllll|}
\hline
{\bf System}                         & {\bf LR\%}    & {\bf LP\%}    & {\bf CB}     & {\bf 0CB\%}     & {\bf 2CB\%} \\
\hline
Minimal~\cite{Charniak97}              & 83.4    & 84.1    & 1.40    & 53.2      & 79.0  \\
Magerman95~\cite{Magerman95}           & 84.6    & 84.9    & 1.26    & 56.6      & 81.4  \\
Charniak97~\cite{Charniak97}           & 87.5    & 87.4    & 1.00    & 62.1      & 86.1  \\
Collins97~\cite{Collins97}             & 88.1    & 88.6    & 0.91    & 66.4      & 86.9  \\
Charniak99~\cite{Charniak99}           & 90.1    & 90.1    & 0.74    & 70.1      & 89.6  \\
\hline
SCFG~\cite{Charniak97}                 & 71.7          &      75.8     &      2.03    &       39.5      &       68.1 \\
{\sl T-gram $(d\leq 5~(2^{PH}))$}          & {\sl 82.9}    & {\sl 85.1}    & {\sl 1.30}    & {\sl 58.0}      & {\sl 82.1} \\
\hline
\end{tabular}
}
\caption{Various results on WSJ section~23 sentences $\leq~40$ words (2245 sentences).}
\label{FigResults}
\end{table*}
}
First we review the lexical-conditionings in previous work (other important conditionings are not discussed for 
space reasons). Magerman95~\cite{Magerman95,JelinekEtAl94} grows a decision-tree to estimate $P(T|S)$ through 
a history-based approach which conditions on actual-words. Charniak~\cite{Charniak97} presents lexicalizations 
of SCFGs: the Minimal model conditions SCFG rule generation on the head-word of its left-hand side, while Charniak97 
further conditions the generation of every constituent's head-word on the head-word of its parent-constituent, 
effectively using bilexical dependencies. Collins97~\cite{Collins97} uses a bilexicalized $0^{th}$-order Markov Grammar: 
a lexicalized CFG rule is generated by projecting the head-child first followed by every left and right dependent, 
conditioning these steps on the head-word of the constituent. Collins97 extends this scheme to deal with subcat frames,
adjacency, traces and wh-movement. Charniak99 conditions lexically as Collins does but also exploits up to $3^{rd}$-order 
Markov processes for generating dependents. Except for T-grams and SCFGs, all systems smooth the relative frequencies with 
much care.

Sentences~$\leq 40$ words (including punctuation) in section~23 were parsed by various T-gram systems.
Table~\ref{FigResults} shows the results of some systems including ours. Systems conditioning mostly on lexical
information are contrasted to SCFGs and T-grams. Our result shows that T-grams improve on SCFGs but fall short of 
the best lexical-dependency systems. Being 10-12\% better than SCFGs, comparable with the Minimal model and Magerman95 
and about 7.0\% worse than the best system, it is fair to say that (depth~5) T-grams perform more like bilexicalized 
dependency systems than bare SCFGs. 

\begin{figure}[htb]
\epsfxsize=7.9cm
\center{
\epsfbox{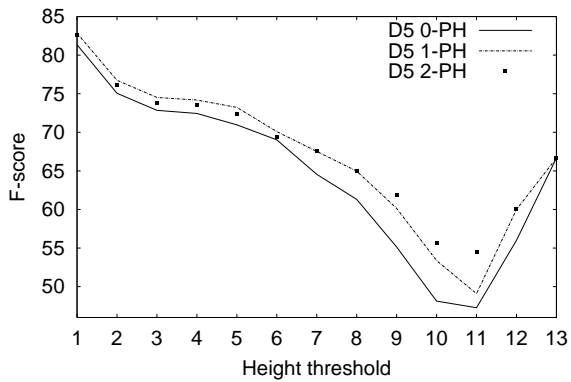}
}
\caption{Heigher nodes are harder}
\label{FigureNodeHeight}
\end{figure}

Table~\ref{TabOfRes} exhibits results of various T-gram systems. Columns~1-2 exhibit the traditional
DOP~observation about the effect of the size of subtrees/T-grams on performance. Columns~3-5 are more 
interesting: they show that even when T-gram size is kept fixed, systems that are pre-head enriched
improve on systems that are not pre-head enriched ($0^{PH}$). This is supported by the result of column~1 in contrast to 
SCFG and Collins97 (table~\ref{FigResults}): the $D1$ T-gram system differs from Collins97 almost only in 
pre-head vs.\ head enrichment and indeed performs midway between SCFG and Collins97.
This all suggests that allowing bilexical dependencies in T-gram models should improve performance. 
It is noteworthy that pre-head enriched systems are also more efficient in time and space.
Column~6 shows that adding Markovian conditioning to subcat frames further improves performance suggesting
that further study of the conditional probabilities of dependent T-grams is necessary.
Now for any node in a gold / proposed parse, let node-height be the average path-length to a word dominated by that node. 
We set a threshold on node-height in the gold and proposed parses and observe performance. Figure~\ref{FigureNodeHeight} 
plots the F-score = (2*LP*LR)/(LP+LR) against node-height threshold. Clearly, performance degrades as the nodes get further 
from the words while pre-heads improve performance.
{
{ 
\Xsmall
\begin{table*}[tbh]
\begin{tabular}{|c|cc|ccc|c|}
\hline
{\bf SYSTEM}       & $D^1(2^{PH})$  	& $D^4(2^{PH})$ & $D^5(2^{PH})$ & $D^5(1^{PH})$ & $D^5(0^{PH})$ & $D^5(2^{PH})+Markov$ \\
\hline
LR           & 80.03    	& 82.42         &  82.57       &   82.85        &   81.35    &   82.93               \\
LP           & 80.99    	& 85.23  	&  85.02       &   85.06        &   84.59    &   85.13               \\
CB           & 1.70     	& 1.32   	&  1.44        &   1.43         &   1.48     &   1.30                \\
\#sens       & \multicolumn{2}{c|}{2245}            & \multicolumn{3}{c|}{first 1000}                      &  2245 \\
\hline
\end{tabular}
\caption{Results of various systems: $D^i$ ($d\leq i$), $i^{PH}$ (pre-head length is $i$), $+Markov$ ($1^{st}$ order Markov conditioning on 
nodes with empty subcat frames for generating $L$ and $R$ T-grams).}
\label{TabOfRes}
\end{table*}
}
}

\section{Conclusions}
\label{SecConcs}
We started this paper wondering about the merits of structural-relations. We presented
the T-gram model and exhibited empirical evidence for the usefulness as well as the shortcomings
of structural relations. We also provided evidence for the gains from enrichment 
of structural relations with semi-lexical information. In our quest for better modeling, we still
need to explore how structural-relations and bilexical dependencies can be combined.
Probability estimation, smoothing and efficient implementations need special attention.

{\bf Acknowledgements:}
I thank Remko Scha, Remko Bonnema,
Walter Daelemans and Jakub Zavrel for illuminating discussions,  Remko
Bonnema for his pre- and post-parsing software, and
the anonymous reviewers for their comments. This work is supported
by the Netherlands Organization for Scientific Research.

\bibliography{BIBLIOGRAPHY}
\end{document}